\documentclass[10pt,twocolumn,letterpaper]{article}

\usepackage{cvpr}              

\usepackage{pgfplots}

\definecolor{cvprblue}{rgb}{0.21,0.49,0.74}
\usepackage[pagebackref,breaklinks,colorlinks,allcolors=cvprblue]{hyperref}

\usepackage[accsupp]{axessibility} 

\def\paperID{16} 
\def\confName{CVPR}
\def\confYear{2026}

\title{SAMIDARE: Advanced Tracking-by-Segmentation for Dense Scenarios}

\author{Shozaburo Hirano and Norimichi Ukita\\
Toyota Technological Institute\\
{\tt\small \{sd25435,ukita\}@toyota-ti.ac.jp}
}

\begin{document}
\maketitle
\begin{abstract}
Automated sports analysis demands robust multi-object tracking (MOT), yet segmentation-based methods often struggle with mask errors and ID switches in dense scenes. We propose SAMIDARE, a framework that enhances SAM2MOT for crowded scenes through three key components: (1) density-aware mask re-generation and (2) selective memory updates, both for adaptive mask control to preserve target feature integrity, and (3) state-aware association and new track initialization, which improves robustness under mutual occlusions and frequent frame-out events. Evaluated on the SportsMOT dataset, SAMIDARE achieves state-of-the-art performance, outperforming the baseline by 2.5 HOTA and 4.2 IDF1 points on the validation set. These results demonstrate that adaptive feature management using mask control and state-aware association provide a robust and efficient solution for dense sports tracking. Code is available at https://github.com/ZabuZabuZabu/SAMIDARE

\end{abstract}    
\section{Introduction}
\label{sec:intro}

\begin{figure}
    \centering
    \begin{subfigure}{\linewidth}
        \centering
        \includegraphics[width=0.85\linewidth]{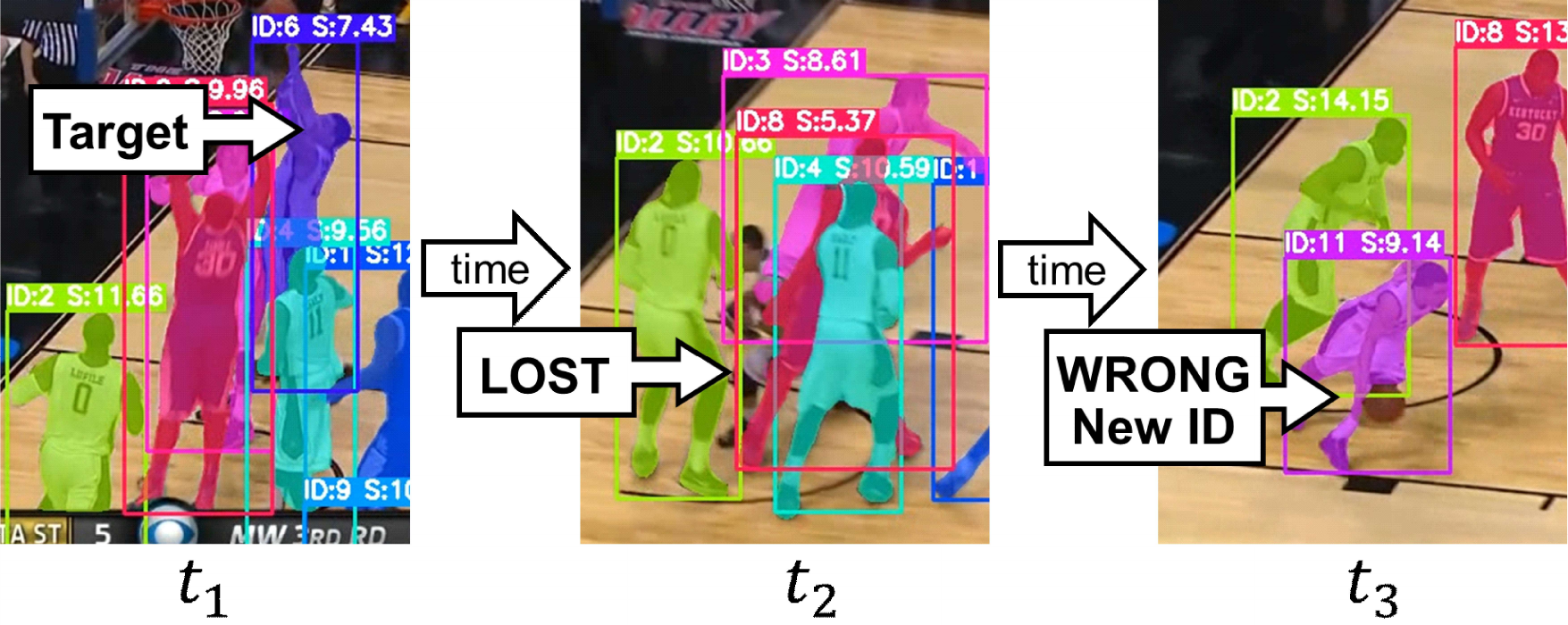}
        \caption{SAM2MOT}
        \label{subfig:first_fig1}
    \end{subfigure}
    
    \vspace{1em} 
    
    \begin{subfigure}{\linewidth}
        \centering
        \includegraphics[width=0.85\linewidth]{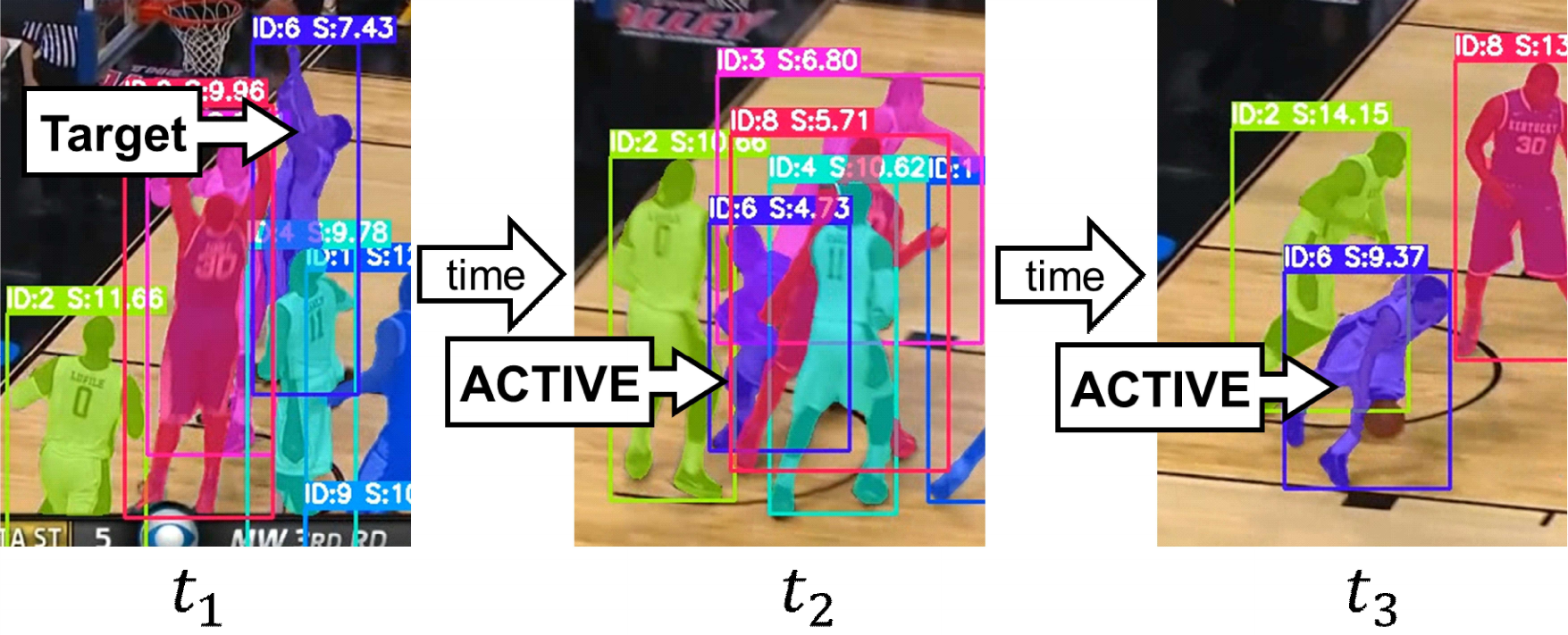}
        \caption{SAMIDARE}
        \label{subfig:first_fig2}
    \end{subfigure}
    
    \caption{Comparison in a dense basketball scenario. In SAM2MOT (top), Player proximity results in target loss (LOST) in $t_{2}$ because of an inappropriate mask update under a heavily crowded scene in $t_{1}$. This failure leads to the wrong new ID assignment (WRONG New ID) in $t_{3}$. SAMIDARE (bottom) successfully maintains continuous tracking (ACTIVE) through adaptive mask control.}
    \label{fig:first_fig}
\end{figure}

Multi-object tracking (MOT) in sports broadcasting is challenging due to rapid camera motion, fast player motions, and frequent occlusions. While tracking-by-detection~\cite{DBLP:conf/eccv/ZhangSJYWYLLW22,DBLP:conf/iccv/CuiZZYWW23,DBLP:conf/cvpr/CaoPWKK23,DBLP:conf/wacv/HuangYSKKLHH24} remains dominant, its reliance on detector quality often causes fragmented trajectories and ID switches in dense scenes with heavy player overlap.

To address these limitations in crowded scenes, tracking-by-segmentation~\cite{DBLP:conf/cvpr/LiKDPSG024,DBLP:journals/tip/YangHCJH26,DBLP:journals/corr/abs-2410-16268} shifts the focus toward temporal mask propagation rather than associating object bounding boxes (BBoxes). SAM2MOT~\cite{DBLP:journals/corr/abs-2504-04519} exploits the memory bank of SAM2~\cite{DBLP:conf/iclr/RaviGHHR0KRRGMP25} for mask-based tracking. Using masks can reduce the influence of features from non-target objects and background regions, thereby improving robustness in crowded scenes. However, in highly dense crowded scenes, mask expansion or drift to neighboring objects still persists. Moreover, existing memory management mechanisms remain insufficient, leading to lost tracks and ID switches (Fig.~\ref{fig:first_fig}). 

We therefore propose SAMIDARE (SAM2MOT Improved for Dense AREa), which enhances memory management in crowded sports scenes, as shown in Fig.~\ref{fig:first_fig}. In addition, SAMIDARE improves data association for track continuation and re-activation in severe crowds. Our main contributions are as follows:

\begin{figure}[t]
    \centering
    \begin{subfigure}{\linewidth}
        \centering
        \includegraphics[width=\linewidth]{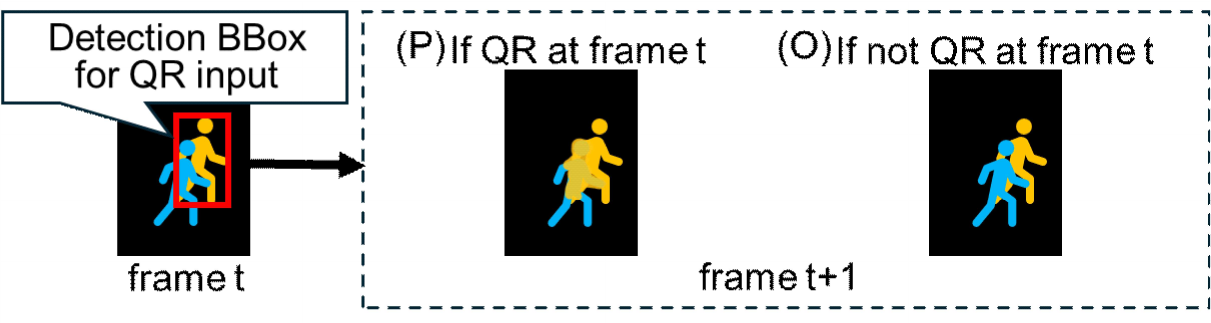}
        \caption{Regulating the re-generation of masks in crowded scenes.}
        \label{subfig:case1}
    \end{subfigure}
    
    \vspace{1em} 
    
    \begin{subfigure}{\linewidth}
        \centering
        \includegraphics[width=\linewidth]{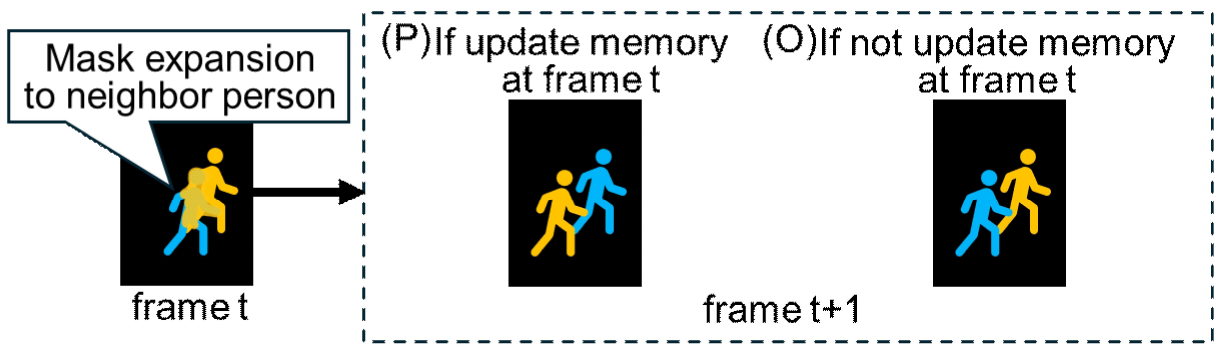}
        \caption{Identifying the person whose mask is expanding by hybrid evaluation.}
        \label{subfig:case2}
    \end{subfigure}
    
    \vspace{1em}
    
    \begin{subfigure}{\linewidth}
        \centering
        \includegraphics[width=\linewidth]{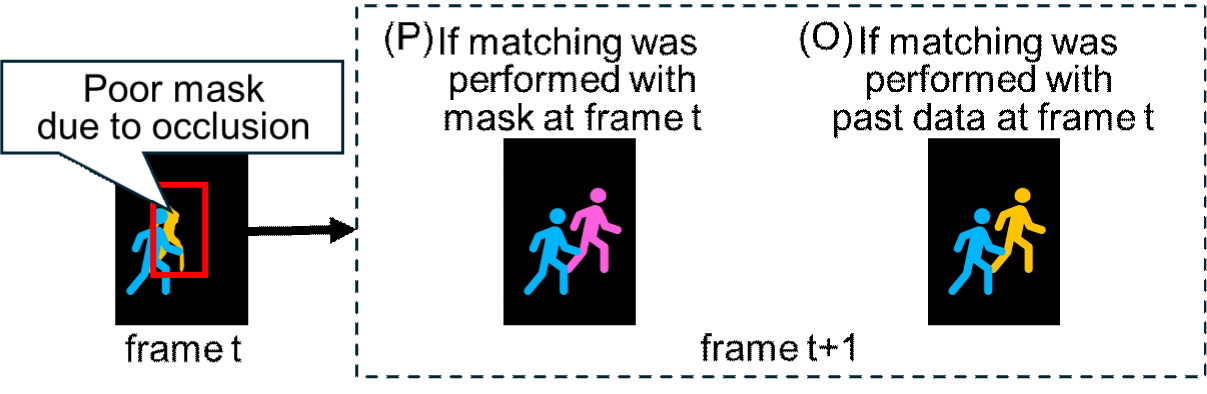}
        \caption{Matching for the track whose mask quality is low.}
        \label{subfig:case3}
    \end{subfigure}
    
    \caption{Contributions of our method. Here, only focus on the person with the yellow mask for a brief explanation. The blue mask person is displayed as an occlusion distractor. Red boxes indicate the target's detected bounding boxes. In each case \ref{subfig:case1}, \ref{subfig:case2}, \ref{subfig:case3}, the leftmost image represents the mask state at frame $t$, while the center and rightmost images represent two patterns of mask states at frame $t+\alpha$.}
    \label{fig:contribution}
\end{figure}

\textbf{1. Density-Aware Quality Reconstruction (DA-QR)}:
Although SAM2MOT uses Quality Reconstruction (QR) to recover degraded masks with a detection BBox (red box in Fig.~\ref{subfig:case1}), QR can fail in crowded scenes because the BBox may include nearby players (blue mask at $t$ in Fig.~\ref{subfig:case1}). This often causes mask drift or expansion at later frames ((P) at $t+1$ in Fig.~\ref{subfig:case1}). DA-QR addresses this issue by suppressing QR in crowded regions ($t$ in Fig.~\ref{subfig:case1}) and enabling it when the target becomes more isolated, thereby preventing erroneous regeneration ((O) at $t+1$ in Fig.~\ref{subfig:case1}).

\textbf{2. Hybrid Cross-Object Interaction (H-CoI)}:
SAM2MOT uses Cross-object Interaction (CoI) to avoid memory updates from unreliable masks. However, because it mainly relies on the variance of mask confidence scores, it can miss stable but low-quality masks (yellow mask at $t$ in Fig.~\ref{subfig:case2}), allowing features from nearby objects (blue mask at $t$ in Fig.~\ref{subfig:case2}) to corrupt the memory and cause later ID switches ((P) at $t+1$ in Fig.~\ref{subfig:case2}). H-CoI instead uses both the mean and variance of mask confidence scores to identify unreliable masks more robustly, preventing such errors (yellow mask in (O) at $t+1$ in Fig.~\ref{subfig:case2}).

\textbf{3. State-Aware Object Addition (SA-OA)}:
When an unmatched detection BBox (red box in Fig.~\ref{subfig:case3}) corresponds to an inactive track, SAM2MOT generates its mask using stored memory (yellow mask in Fig.~\ref{subfig:case3}). However, after occlusion or out-of-frame periods, the recovered mask can be inaccurate, which may cause incorrect new track initialization (pink mask in (P) in Fig.~\ref{subfig:case3}). SA-OA addresses this issue by additionally using the inactive track's BBox, whose location and scale remain more stable, thereby improving re-activation with the correct historical ID (yellow mask in (O) at $t+1$ in Fig.~\ref{subfig:case3}).
\section{Related work}
\label{sec:related_work}

\begin{figure*}
    \centering
    \includegraphics[width=\linewidth]{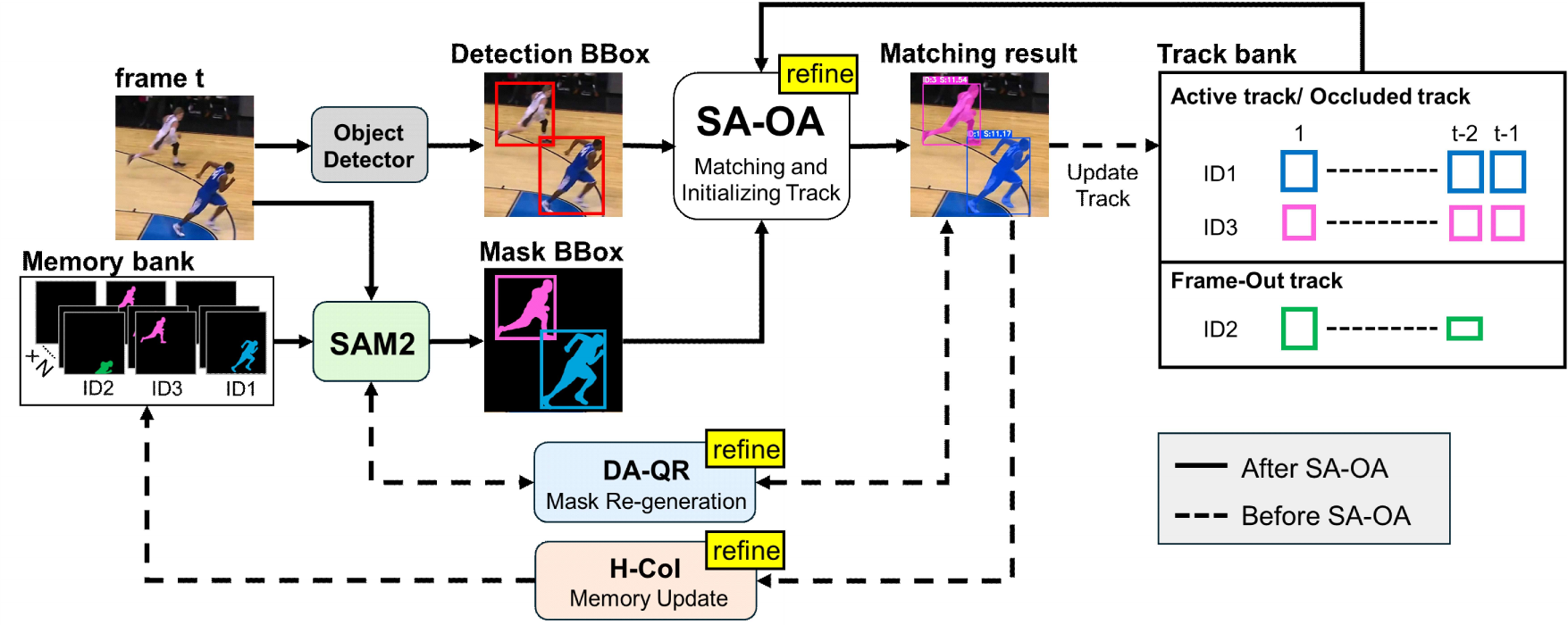}
    \caption{Overview of the SAMIDARE pipeline. Our framework enhances mask propagation by integrating adaptive mask control (DA-QR and H-CoI) and a three-stage matching strategy. Solid and dashed arrows represent processing before and after matching, respectively. Modules labeled “refine” refer to modules that refine the conventional method.}
    \label{fig:overview}
\end{figure*}

\subsection{Tracking-by-Detection}

Tracking-by-Detection is the dominant paradigm in multi-object tracking (MOT). It detects bounding boxes (BBoxes) in each frame and associates them over time using motion models or appearance features~\cite{DBLP:journals/bmcbi/VeeramaniRC18,DBLP:journals/tmm/DuZSZSGM23,DBLP:journals/corr/abs-2206-14651,DBLP:conf/cvpr/QinZ0D0T23}. Representative methods such as ByteTrack~\cite{DBLP:conf/eccv/ZhangSJYWYLLW22} and DeepEIoU~\cite{DBLP:conf/wacv/HuangYSKKLHH24} use IoU- or ReID-based cues for data association.

However, because these methods rely on rectangular BBoxes, they often capture features from neighboring objects in crowded scenes. This feature contamination makes object association less reliable.

\subsection{Tracking-by-Segmentation}

Tracking-by-Segmentation is an emerging approach that focuses on the temporal propagation of pixel-level masks. By using segmentations instead of BBoxes, target features are less likely to be contaminated by neighboring objects, which improves association accuracy in dense scenarios. A notable example is SAM2MOT~\cite{DBLP:journals/corr/abs-2504-04519}, which adapts the memory bank of SAM2~\cite{DBLP:conf/iclr/RaviGHHR0KRRGMP25} for MOT and enables mask-based target representation over time.

However, SAM2 still suffers from mask expansion and mask drift toward nearby objects in high-density environments containing similar objects in close proximity, such as sports scenes. When such mask errors occur, features from other objects are introduced into the target memory, which hinders stable tracking and degrades target feature integrity. Although SAM2MOT introduces memory-management modules to mitigate memory corruption caused by mask errors, these modules are still insufficient for highly dense sports scenes.

\section{Proposed Method}
\label{sec:proposed_method}

\subsection{Overview}

Figure~\ref{fig:overview} shows the overview of our proposed framework, SAMIDARE. Built on tracking-by-segmentation using SAM2MOT, SAMIDARE refines key modules (``refine'' in Fig.~\ref{fig:overview}) for dense scenes to improve ID consistency through the following frame-by-frame process:

\begin{enumerate}
    \item \textbf{Initialization}: In the first frame, detection BBoxes from an object detector (e.g., YOLO-X~\cite{DBLP:journals/corr/abs-2107-08430}) are used as BBox prompts for SAM2 to initialize tracks.
    \item \textbf{Mask Propagation}: In each subsequent frame, SAM2 (light green module in Fig.~\ref{fig:overview}) predicts masks and their confidence scores for all tracks using the current frame and the memory bank. These masks are then passed to SA-OA (white module in Fig.~\ref{fig:overview}).
    \item \textbf{Object Detection}: The detector (gray module in Fig.~\ref{fig:overview}) detects BBoxes from the current frame and sends them to SA-OA.
    \item \textbf{State-Aware Object Addition (SA-OA) (Sec.~\ref{subsection:SA-OA})}: SA-OA matches detection BBoxes and the masks depending on track states such as occlusion and out-of-frame, enabling robust track recovery and new ID assignment.
    \item \textbf{Adaptive Mask Control}: After SA-OA, SAMIDARE preserves the reliable memory bank through two modules:
    \begin{itemize}
        \item \textbf{Density-Aware Quality Reconstruction (DA-QR) (Sec.~\ref{subsection:DA-QR})}: DA-QR suppresses mask re-generation in crowded regions to avoid memory corruption caused by overlapping objects.
        \item \textbf{Hybrid Cross-object Interaction (H-CoI) (Sec.~\ref{subsection:H-COI})}: H-CoI (orange module in Fig.~\ref{fig:overview}) skips memory updates for unreliable masks to prevent corrupted memory from degrading future tracking.
    \end{itemize}
\end{enumerate}

\subsection{Density-Aware Quality Reconstruction}
\label{subsection:DA-QR}

SAM2MOT employs Quality Reconstruction (QR) to improve long-term tracking stability by re-generating masks when their quality begins to degrade. QR is triggered when the mask confidence score of the $i$-th target, denoted by $S_{\text{mask},i}$, falls between $\tau_r$ and $\tau_p$. This criterion is designed to trigger reconstruction before severe mask degradation occurs. Specifically, QR uses, as a new SAM2 prompt, the detection BBox matched to the degraded mask in Stage 1 matching (Sec.~\ref{subsection:SA-OA}). However, QR can fail in crowded scenes: the prompted BBox may include neighboring people, causing the mask to expand into neighboring objects in subsequent frames.

To mitigate this issue, we introduce Density-Aware QR (DA-QR), which regulates mask re-generation in QR based on local density. We quantify this local density $\text{D}(T_i)$ for a target track $T_i$ using its matched detection BBox $B_{\text{det},i}$. $\text{D}(T_i)$ is calculated by summing the intersection ratios between $B_{\text{det},i}$ and all other detection BBoxes $B_{\text{det},j}$ ($j \neq i$) present in the frame, where $j$ indexes all other detected people in the frame:
\begin{equation}
\text{D}(T_i) = \sum_{j \neq i} \frac{\text{Area}(B_{\text{det},i} \cap B_{\text{det},j})}{\text{Area}(B_{\text{det},i})} \nonumber
\end{equation}
where $\text{Area}(\cdot)$ denotes the area in pixels. Note that $D(T_i)$ is not normalized to $[0,1]$; it accumulates overlaps with all other detection BBoxes.
In DA-QR, QR is executed when the mask confidence score $S_{\text{mask},i}$ falls between $\tau_r$ and $\tau_p$, as in SAM2MOT, and the following condition is satisfied:
\begin{equation}
\text{D}(T_i) < \theta_{\text{density}}
\label{eq:DA-QR}
\end{equation}
This adaptive mask re-generation maintains the purity of the target's memory, improving long-term ID consistency throughout and after dense situations.

\subsection{Hybrid Cross-object Interaction}
\label{subsection:H-COI}

SAM2MOT employs Cross-object Interaction (CoI), which identifies unreliable masks (e.g., erroneous mask expansion due to overlapping people) to prevent memory corruption by avoiding memory updates for such unreliable overlapping masks.
CoI has two steps.

{\bf Step 1}: To detect overlap of masks, CoI calculates the mask Intersection-over-Union (mIoU) for every pair of tracked persons' masks, $M_i$ and $M_j$, in the current frame:
\begin{equation}
\text{mIoU}(M_i, M_j) = \frac{|M_i \cap M_j|}{|M_i \cup M_j|}
\nonumber
\end{equation}

{\bf Step 2}: When $\text{mIoU}(M_i, M_j) > \theta_{\text{miou}}$, we compare the variances of the mask-confidence scores over the recent $N$ frames for the two masks, $\sigma_i^{2}$ and $\sigma_j^{2}$. We then regard the mask with the larger variance as unreliable (i.e., $M_i$ if $\sigma_i^{2}>\sigma_j^{2}$, otherwise $M_j$).
This logic assumes that high variance indicates mask segmentation instability. CoI skips memory updates for such unreliable masks to prevent memory corruption. 

However, a temporally stable, unreliable mask may have low variance with low scores. For such a track, the memory is updated inappropriately so that it includes others’ mask features, resulting in an ID switch in later frames.

We therefore replace CoI with our proposed Hybrid Cross-Object Interaction (H-CoI) to identify which mask is more unreliable by adaptively selecting the most discriminative indicator between the mean $\mu$ and variance $\sigma^2$.
Step 1 is shared between CoI and H-CoI. In Step 2 of H-CoI, given $\Delta \mu = |\mu_i - \mu_j|$ and $\Delta \sigma^2 = |\sigma_i^2 - \sigma_j^2|$, the unreliable mask is identified as follows:
\begin{equation}
\begin{split}
\arg\min_{k \in \{i,j\}} \mu_k
&\quad \text{if } \Delta \mu \ge \Delta \sigma^2 \\
\arg\max_{k \in \{i,j\}} \sigma_k^2
&\quad \text{if } \Delta \mu < \Delta \sigma^2
\end{split}
\label{eq:H-CoI}
\end{equation}
Once a mask and its track (either $i$ or $j$) are selected as unreliable, H-CoI skips its memory update in the current frame.

\subsection{State-Aware Object Addition (SA-OA)}
\label{subsection:SA-OA}

In SAM2MOT, the Object Addition module performs Hungarian matching between detection BBoxes $\{ B_{\text{det},i} \}$ and mask BBoxes $\{ B_{\text{mask},j} \}$, each of which is the minimum enclosing boxes of the masks, of all tracks in the current frame to find newly appeared people. Each unmatched detection BBox with a high confidence score is regarded as a new person instance and fed into SAM2 as a prompt to initialize the mask of the track for this person. A new ID is assigned to the new person. 

Note that the above matching process in the current frame, $t$, is performed not only for the mask BBox of an active track but also for that of an inactive track, which is not matched to any detection BBox in $t-1$; SAM2MOT maintains inactive tracks during several frames (i.e., 60 frames in the SAM2MOT implementation) after track inactivation (e.g., due to occlusion or leaving the field of view). If matching is successful for any inactive track, it is re-activated as an active track.

However, in such a re-activation time, inactive tracks often suffer lower mask-shape confidence due to appearance shifts occurring (e.g., changes in whole-body orientation and pose) while they are not detected/observed. Such a poorly segmented mask is difficult to re-associate with detection BBoxes due to low IoU.

To address this limitation, we introduce State-Aware Object Addition (SA-OA), a hierarchical matching that considers the following three track states:

\begin{itemize}
    \item \textbf{Active tracks}: Tracks matched in frame $t-1$.
    \item \textbf{Occluded tracks}: Tracks that are unmatched in $t-1$ but exhibited a high density $\text{D}(T_i) > \delta$ in their last matched frame. A high $\text{D}(T_i)$ indicates that the person was occluded by other people.
    \item \textbf{Frame-Out tracks}: Tracks that were unmatched in $t-1$ and maintained a low density $\text{D}(T_i) \le \delta$ in their last matched frame,
    implying that these tracks may leave the field of view.
\end{itemize}
Based on these three states, SA-OA performs a three-stage matching for new ID assignment:

\subsubsection{Stage 1: Matching for Active and Occluded tracks}
Stage 1 associates detection BBoxes $B_{\text{det},j}$ with Active and Occluded tracks $T_i$ by applying the Hungarian algorithm to a cost matrix $\text{C}(T_i, B_{\text{det},j})$ defined by the following function:
\begin{equation}
\text{C}(T_i, B_{\text{det},j}) = w (1 - \text{IoU}(B_{\text{mask},i},B_{\text{det},j})) + (1 - w)(1 - \mu_i)
\label{eq:cost}
\end{equation}
where
$\mu_i$ is the mean of the mask confidence score for track $T_i$. $w$ is a weight balancing the spatial similarity term ($\text{IoU}$) and mask reliability term ($\mu_i$).

This formulation prioritizes reliable masks and reduces mismatches caused by unreliable masks.

\subsubsection{Stage 2: Matching for inaccurate masks}
Stage 2 targets Active and Occluded tracks that failed to match in Stage 1. This failure means that tracks remain unmatched despite the presence of unmatched detection BBoxes. One reason for this failure is that the mask for each remaining track is inaccurately generated (e.g., fragmentation or shifting toward adjacent people) due to occlusion, resulting in a low IoU with detection BBoxes that prevents successful matching. Therefore, we associate these tracks using their last matched detection BBox $B_{\text{last},i}$ instead of the inaccurate masks. If matched, QR is applied to reset the inaccurate masks.

\subsubsection{Stage 3: Matching for Frame-Out tracks}
Stage 3 addresses Frame-Out tracks, which are likely to leave the field of view temporarily. Continuing mask generation for such tracks may cause SAM2 to generate masks for visually similar people within the frame. Moreover, in sports scenarios where frame-out events frequently occur, maintaining mask generation for Frame-Out tracks incurs unnecessary computational overhead.
Therefore, we avoid mask generation for Frame-Out tracks and instead compare the latest detection BBoxes of these tracks observed in $t-\alpha$ with the remaining BBoxes detected in $t$ based on spatial continuity in the track positions.
Only detections unmatched after all three stages are initialized as new entries.

\section{Experiments and discussion}
\label{sec:experiments}

\begin{table}
\centering
\caption{Comparing SAMIDARE with state-of-the-art tracking-by-detection methods on the SportsMOT test set. Methods marked with $^*$ indicate that modules other than the detector (e.g., motion or appearance models) are trained or fine-tuned on the SportsMOT dataset.}
\label{tab:TBD}
\begin{tabular}{lccc}
\toprule
Method & HOTA$\uparrow$ & IDF1$\uparrow$ \\ \hline \hline
ByteTrack\cite{DBLP:conf/eccv/ZhangSJYWYLLW22}  & 64.1 & 71.4  \\
MixSort-Byte$^*$\cite{DBLP:conf/iccv/CuiZZYWW23}  & 65.7 & 74.1  \\
OC-SORT\cite{DBLP:conf/cvpr/CaoPWKK23} & 73.7 & 74.0  \\
MixSort-OC$^*$\cite{DBLP:conf/iccv/CuiZZYWW23}  & 74.1 & 74.4 \\
GeneralTrack\cite{DBLP:conf/cvpr/Qin0ZF0024}  & 74.1 & 76.4  \\
DeepEIoU$^*$\cite{DBLP:conf/wacv/HuangYSKKLHH24} & 75.8 & 77.6 \\
DiffMOT$^*$\cite{DBLP:conf/cvpr/LvHZLH024}  & 76.2 & 76.1  \\
McByte\cite{DBLP:conf/cvpr/StanczykYB25} &76.9 & 77.5\\ \hline \hline
SAM2MOT\cite{DBLP:journals/corr/abs-2504-04519} & 76.4 & 76.5  \\
\textbf{ours} & \textbf{77.3} & \textbf{78.6}  \\
\bottomrule
\end{tabular}
\end{table}

\begin{table}
\centering
\caption{Comparing SAMIDARE with the other tracking methods using segmentation mask on the SportsMOT validation set }
\label{table:mask}
\begin{tabular}{lccc}
\hline
Method & HOTA $\uparrow$ & IDF1 $\uparrow$  \\ \hline \hline
DEVA\cite{DBLP:conf/iccv/ChengOPSL23} & 42.4 & 42.1 \\
Grounded SAM 2\cite{DBLP:conf/eccv/LiuZRLZYJLYSZZ24}\cite{DBLP:conf/iccv/KirillovMRMRGXW23} & 66.1 & 70.2  \\
MASA \cite{DBLP:conf/cvpr/LiKDPSG024}& 73.6 & 71.2  \\
McByte \cite{DBLP:conf/cvpr/StanczykYB25}& 83.9 & 83.6  \\ \hline \hline
SAM2MOT \cite{DBLP:journals/corr/abs-2504-04519}& 83.7 & 82.9  \\ 
\textbf{ours} & \textbf{86.2} & \textbf{87.1}\\ \hline
\end{tabular}
\end{table}

\subsection{Dataset and Evaluation Metrics}
We evaluate SAMIDARE on the SportsMOT dataset~\cite{DBLP:conf/iccv/CuiZZYWW23}, which consists of 240 video sequences covering basketball, football, and volleyball.
To assess performance, we use Higher Order Tracking Accuracy~(HOTA)~\cite{DBLP:journals/ijcv/LuitenODTGLL21} as our primary metric to evaluate the balance between detection and association. The Identity F1 Score (IDF1)~\cite{DBLP:conf/eccv/RistaniSZCT16} is also used to measure the quality of identity preservation over time.

\subsection{Implementation details}

We use the authors' implementation of SAM2MOT and its hyperparameters, available at~\cite{DBLP:journals/corr/abs-2504-04519}.
With this implementation, SAMIDARE is implemented by (i) adding the additional QR trigger defined by Eq.~\ref{eq:DA-QR}, (ii) replacing CoI with H-CoI defined by Eq.~\ref{eq:H-CoI}, and (iii) adding two stages for Object Addition (i.e., Stage 2 and Stage 3).
In addition, SAM2 in this implementation is replaced by SAM 2.1 ~\cite{sam2github} (Segment Anything Model 2.1) as the core engine for pixel-level object segmentation and temporal mask propagation. Specifically, we employ the Hiera-Large version of the pre-trained SAM 2.1 model to ensure high-fidelity mask generation. 
Note that, while these SAM2MOT and SAM2.1 used in our experiments are trained on datasets of various domains, no sports dataset is used for training. Experiments using these SAM2MOT and SAM2.1, which are not optimized for sports videos, are suitable for evaluating the generalizability of our proposed method.
For the object detection, on the other hand, we adopt the YOLOX-X detector~\cite{DBLP:journals/corr/abs-2107-08430}, which is pre-trained on the SportsMOT~\cite{DBLP:conf/iccv/CuiZZYWW23} provided in ~\cite{DBLP:conf/cvpr/StanczykYB25}. This YOLO-X detector is widely used in many previous experiments~\cite{DBLP:conf/cvpr/StanczykYB25,DBLP:conf/iccv/CuiZZYWW23,DBLP:conf/cvpr/LvHZLH024} in the literature of sport video analysis. Therefore, to ensure a fair and direct comparison with the baseline, SAM2MOT, and other state-of-the-art methods, we use the same detector for all experiments.

The hyperparameters for our proposed modules are fixed across all experiments to demonstrate the generalizability of SAMIDARE without the need for per-video tuning. The density threshold for DA-QR in (Eq.~\ref{eq:DA-QR}) is fixed at $\theta_{\text{density}}=1.5$. For SA-OA, we set the threshold for categorizing Occluded tracks and Frame-Out tracks at $\delta=0.6$, and the balancing weight for the cost function (Eq.~\ref{eq:cost}) is set to $w=0.5$. These hyperparameters are set to the parameters that yielded the best accuracy on the validation data of SportsMOT.

\subsection{Comparison with state-of-the-art methods}

We evaluate SAMIDARE against various state-of-the-art trackers.
As shown in Tab.~\ref{tab:TBD}, our method achieves superior results on the test set, outperforming representative Tracking-by-Detection methods such as ByteTrack and DeepEIoU. Notably, SAMIDARE surpasses McByte, the previous top-performing method, in both HOTA and IDF1 metrics. This gain is obtained by SAMIDARE's robustness for crowded scenes.

The effectiveness of SAMIDARE is further highlighted in Tab.~\ref{table:mask}, which provides a comparison with recent mask-based trackers on the validation set. SAMIDARE outperforms the baseline, SAM2MOT, improving HOTA by 2.5\% and IDF1 by 4.2\%. These results suggest that in scenes where visually similar people heavily overlap, such as sports, mask-based tracking requires explicit strategies to handle crowding, as implemented in SAMIDARE.

\subsection{Ablation studies}
\label{subsec:ablation}

\begin{table}
\centering
\caption{Ablation study for each module of our proposed method on the SportsMOT validation set. B, V, and S refer to tracking accuracy in basketball, volleyball, and soccer data, respectively.}
\label{table:compo_ablation}
\begin{tabular}{ccccccc}
\hline
DA-QR & H-CoI & SA-OA &  ALL &  B&  V&  S\\ \hline \hline
 & & & 83.7 & 86.4 & 87.3 & 79.8 \\
\checkmark & & & 83.9 & 86.9 & 86.6 & 80.0 \\
&\checkmark & & 83.2 & 85.9 & 86.8 & 79.3\\
& &\checkmark & 85.6 & 87.9 & 88.5 & 82.5\\
\checkmark& \checkmark& & 84.0 & 86.9 & 87.0 & 80.0\\
\checkmark & \checkmark & \checkmark & \textbf{86.2} & \textbf{88.5} & \textbf{89.4} & \textbf{82.6} \\ \hline
\end{tabular}
\end{table}

Table~\ref{table:compo_ablation} presents the ablation study for each module, underscoring how our proposed modules address challenges specific to different sports.

\noindent{\bf DA-QR and H-CoI.} 
Our experimental results demonstrate that the Density-Aware Quality Reconstruction (DA-QR) module is particularly effective in high-density scenarios. Specifically, adding DA-QR to the baseline improves the HOTA in Basketball (B) from 86.4\% to 86.9\%, marking a 0.5\% gain. This confirms that DA-QR effectively manages feature contamination in crowded scenes by regulating mask re-generation based on local density. 
Regarding Hybrid Cross-Object Interaction (H-CoI), its standalone introduction decreases overall accuracy to 83.2\%. This performance drop is attributed to the fact that continuous QR in dense scenes often leads to erroneous mask expansion. Once this expanded state persists, SAM2 can continue to output a high confidence score despite the mask's inaccuracy. This deceptively heightened mask confidence score misleads the mean-based identification in H-CoI (i.e., Eq.~\ref{eq:H-CoI}), leading to the update of corrupted mask memory and ID switch.
However, combining H-CoI with DA-QR improves performance to 84.0\%. This complementary effect suggests that DA-QR reduces continuous mask expansion, thereby preventing the output of deceptive mask confidence scores. By maintaining more accurate masks, DA-QR enables effective mean-based identification in H-CoI, as confidence scores become more representative of the actual segmentation quality.

\noindent{\bf SA-OA.} 
The State-Aware Object Addition (SA-OA) module provides the most significant performance boost across all categories. This highlights the effectiveness of our state-aware three-stage matching in mitigating redundant new ID assignments for targets that reappear after occlusion or leaving the frame. The effectiveness of Stage 2 and Stage 3 is discussed in detail in Sec.~\ref{subsubsec:SAOA}.

The fact that the highest overall accuracy of 86.2\% is achieved only when all three modules are integrated confirms that our framework provides a comprehensive solution. Specifically, SAMIDARE achieves superior performance in all categories: 88.5\% in Basketball, 89.4\% in Volleyball, and 82.6\% in Soccer. This success demonstrates that balancing adaptive mask control (DA-QR and H-CoI) with state-aware association (SA-OA) is essential for robust multi-object tracking in crowded scenes.

\subsection{Sensitivity Analysis}

\subsubsection{DA-QR}
\begin{table}
\centering
\caption{Sensitivity analysis of HOTA scores with respect to the density threshold $\theta_{density}$ for DA-QR on the SportsMOT validation set.}
\label{table:daqr_comparison}
\begin{tabular}{ccccccc}
\hline
$\theta_{\text{density}}$ &  ALL &  B&  V&  S\\ \hline \hline
$\theta_{\text{density}}$=0.01 & 85.4 & 87.7 & 87.8 & \textbf{82.8} \\
$\theta_{\text{density}}$=0.5 & 85.5 & 87.9 & 88.1 & 82.6 \\
$\theta_{\text{density}}$=1.0& 86.0 & 88.4 & 89.0 & 82.5 \\
$\theta_{\text{density}}$=1.5& \textbf{86.2} & \textbf{88.5} & \textbf{89.4} & 82.6 \\
$\theta_{\text{density}}$=2.0 & 85.9 & \textbf{88.5} & 88.1 & 82.6\\ \hline
\end{tabular}
\end{table}

\begin{table}
\centering
\caption{Comparison of indicators for H-CoI: variance, mean, and our hybrid indicator on the SportsMOT validation set.}
\label{table:coi_index_comparison}
\begin{tabular}{lcccc}
\hline
indicator &  ALL &  B&  V&  S\\ \hline \hline
var & 84.8 & 86.3 & 88.1 & 82.3 \\
mean & 85.5 & 87.4 & 88.3 & 82.6\\ 
ours & \textbf{86.2} & \textbf{88.5} & \textbf{89.4} & \textbf{82.6} \\ \hline
\end{tabular}
\end{table}

Tab.~\ref{table:daqr_comparison} illustrates the impact of the density threshold $\theta_{\text{density}}$ in (Eq.~\ref{eq:DA-QR}), where the HOTA score peaks at 86.2\% when $\theta_{\text{density}} = 1.5$.
The results reveal that an excessively high threshold (e.g., 2.0) allows masks to be re-generated even during extreme crowds. This leads to the generation of expanding masks, similar to the failures observed in conventional methods. Conversely, an overly restrictive threshold (e.g., 0.5) suppresses the mask re-generation process entirely, failing to maintain high-quality segmentation. These findings confirm that neither unconditional mask re-generation nor uniform restriction is optimal. 

In addition, for basketball and volleyball, the highest tracking accuracy was achieved when $\theta_{\text{density}} = 1.5$. However, for soccer, the highest tracking accuracy was achieved when $\theta_{\text{density}} = 0.01$. This suggests that dynamic threshold adjustment is necessary to achieve higher tracking accuracy due to sport-specific player density and interaction patterns.

\subsubsection{H-CoI}
To validate the adaptive indicator selection in H-CoI, we compare our hybrid approach against using only variance or mean. As shown in Tab.~\ref{table:coi_index_comparison}, the H-CoI (ours) achieves the highest overall accuracy of 86.2\% and consistently outperforms single-indicator approaches across all categories. These results demonstrate that in dense situations where mask confidence scores cannot be fully trusted, dynamically selecting the most discriminative indicator among multiple indicators (e.g., variance and mean) is highly effective for identifying erroneous masks.

\subsubsection{SA-OA}
\label{subsubsec:SAOA}

\begin{table}
\centering
\caption{Effectiveness of each matching stage in SA-OA module on the SportsMOT validation set.}
\label{table:SA-OA}
\begin{tabular}{ccccccc}
\hline
Stage 1 & Stage 2 & Stage 3 &  ALL &  B&  V&  S\\ \hline \hline
\checkmark & & & 82.8 & 85.1 & 86.6 & 79.5 \\
\checkmark&\checkmark & & 83.3 & 85.4 & 87.3 & 80.1\\
\checkmark& &\checkmark & 85.7 & 88.3 & 88.5 & 82.2\\
\checkmark & \checkmark & \checkmark & \textbf{86.2} & \textbf{88.5} & \textbf{89.4} & \textbf{82.6} \\ \hline
\end{tabular}
\end{table}

\begin{figure}
    \centering
    \includegraphics[width=\linewidth]{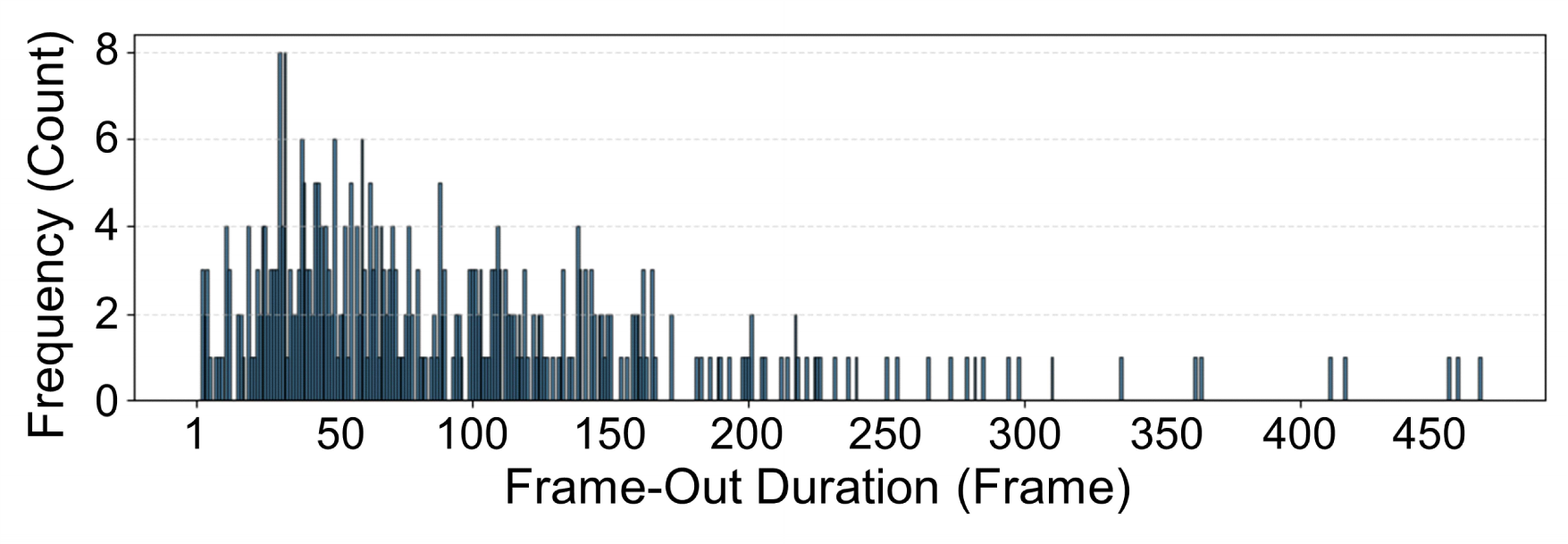}
    \caption{Histogram of frame-out durations on the SportsMOT validation set.}
    \label{fig:frame-out}
\end{figure}

\begin{table}
\centering
\caption{Accuracy and efficiency comparison of re-activation strategies for Frame-Out tracks. We evaluate the performance of mask-based matching (Mask) against matching utilizing the last matched detection BBoxes ($B_{\text{last}}$). "Time" denotes the average processing time for each sport's data.}
\label{table:mask_or_bbox2}
\small
\resizebox{\columnwidth}{!}{
\begin{tabular}{lccccccc}
\toprule
& ALL & \multicolumn{2}{c}{B} & \multicolumn{2}{c}{V} & \multicolumn{2}{c}{S} \\
\cmidrule(lr){2-2} \cmidrule(lr){3-4} \cmidrule(lr){5-6} \cmidrule(lr){7-8}
& HOTA $\uparrow$ & HOTA $\uparrow$ & Time $\downarrow$ & HOTA $\uparrow$ & Time $\downarrow$ & HOTA $\uparrow$ & Time $\downarrow$ \\
\midrule
Mask & 85.1 & 87.8 & 13.1m & 88.3 & 6.0m & 81.0 & 16.5m \\
$B_{\text{last}}$ & \textbf{86.2} & \textbf{88.5} & \textbf{12.8m} & \textbf{89.4} & \textbf{5.6m} & \textbf{82.6} & \textbf{15.1m} \\
\bottomrule
\end{tabular}
}
\end{table}

Tab.~\ref{table:SA-OA} illustrates the contribution of each matching stage within the SA-OA module to the overall tracking accuracy. 

\noindent{\bf Stage 2.} 
The addition of Stage 2 to the baseline matching (Stage 1) results in a steady improvement in HOTA. This result demonstrates that it is more effective to maintain the current trajectory using the last matched detection BBox rather than immediately assigning a new ID for tracks that remain unmatched in Stage 1 due to fragmented masks or masks shifted toward adjacent people after occlusion.

\noindent{\bf Stage 3.} 
The most pronounced performance gain is observed with the introduction of Stage 3. When combined with Stage 1, this stage increases the HOTA in Soccer (S) from 79.5\% to 82.2\% and in Basketball (B) from 85.1\% to 88.3\%. This notable gain underscores the importance of spatial consistency for players re-entering the field of view. The effectiveness of matching using spatial continuity for Frame-Out tracks is supported by the distribution of frame-out durations in the SportsMOT dataset. As shown in Fig.~\ref{fig:frame-out}, the majority of frame-out events last less than 6 seconds (150 frames at 25 fps). In team sports, players generally adhere to specific formations and positions, especially when they are not in proximity to the ball. Consequently, their physical displacement within such a short window remains minimal, enabling Stage 3 matching to successfully re-associate Frame-Out tracks.

To further validate the matching strategy used in Stage 3, we compare the performance of using mask-based matching (the baseline approach in SAM2MOT) against our proposed method using the last matched detection BBox to re-activate Frame-Out tracks. As shown in Tab.~\ref{table:mask_or_bbox2}, utilizing the last matched detection BBoxes ($B_{\text{last}}$) consistently outperforms mask-based matching (Mask) in both tracking accuracy and computational efficiency across all sports. For instance, in the Soccer (S), our approach ($B_{last}$) improves HOTA from 81.0\% to 82.6\% while reducing the average processing time from 16.5m to 15.1m. These results underscore that spatial consistency provided by the historical BBoxes of Frame-Out tracks is a more reliable and computationally efficient cue than mask memory, which often becomes corrupted when a target is unobserved for an extended period.

\subsection{Qualitative evaluation in crowded scenarios}

\begin{figure}
    \centering
    \begin{subfigure}{\linewidth}
        \centering
        \includegraphics[width=\linewidth]{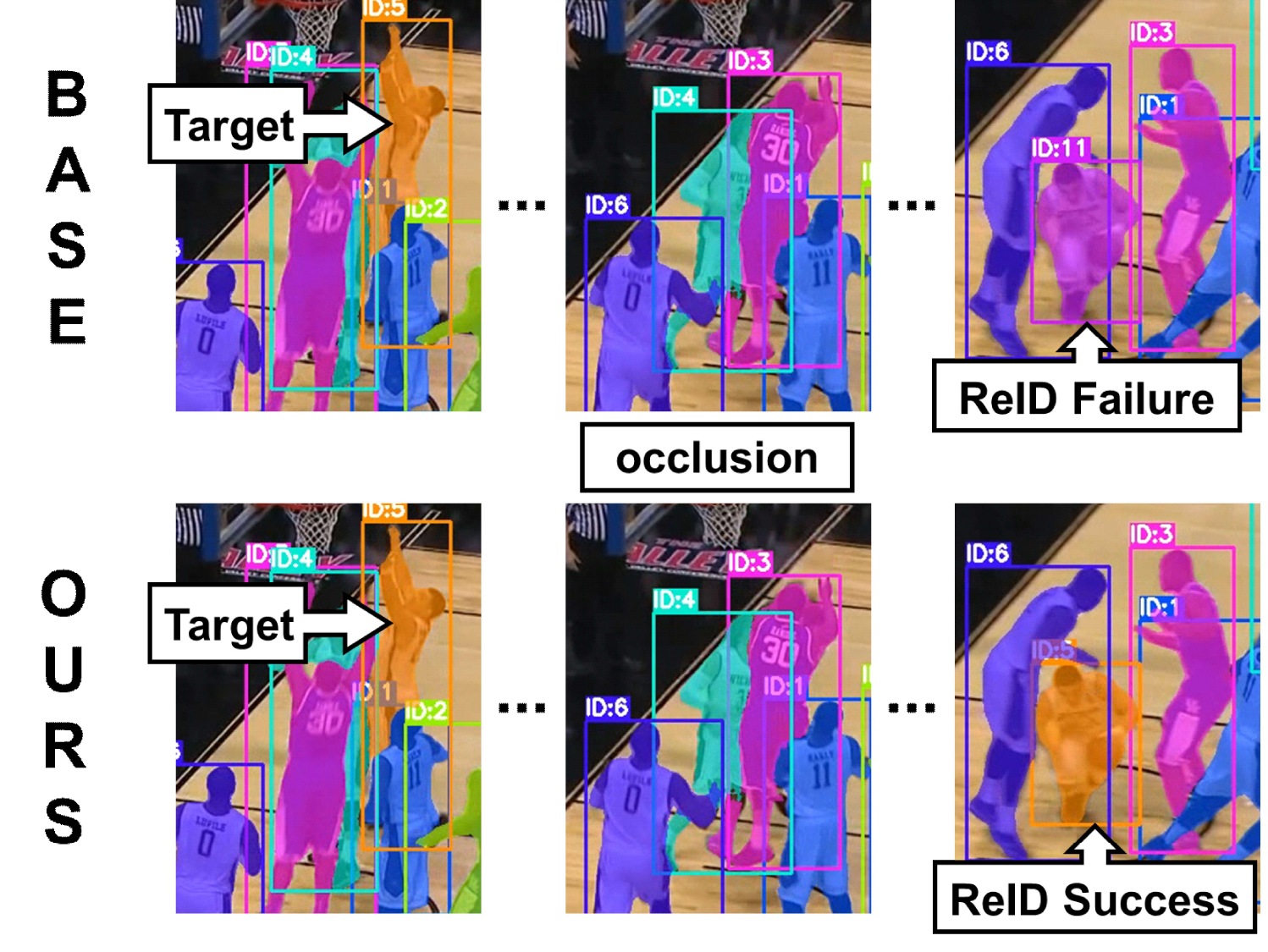}
        \caption{Scene 1: Robustness to crowded situations enabled by DA-QR}
        \label{subfig:result1}
    \end{subfigure}
    
    \vspace{1em} 
    
    \begin{subfigure}{\linewidth}
        \centering
        \includegraphics[width=\linewidth]{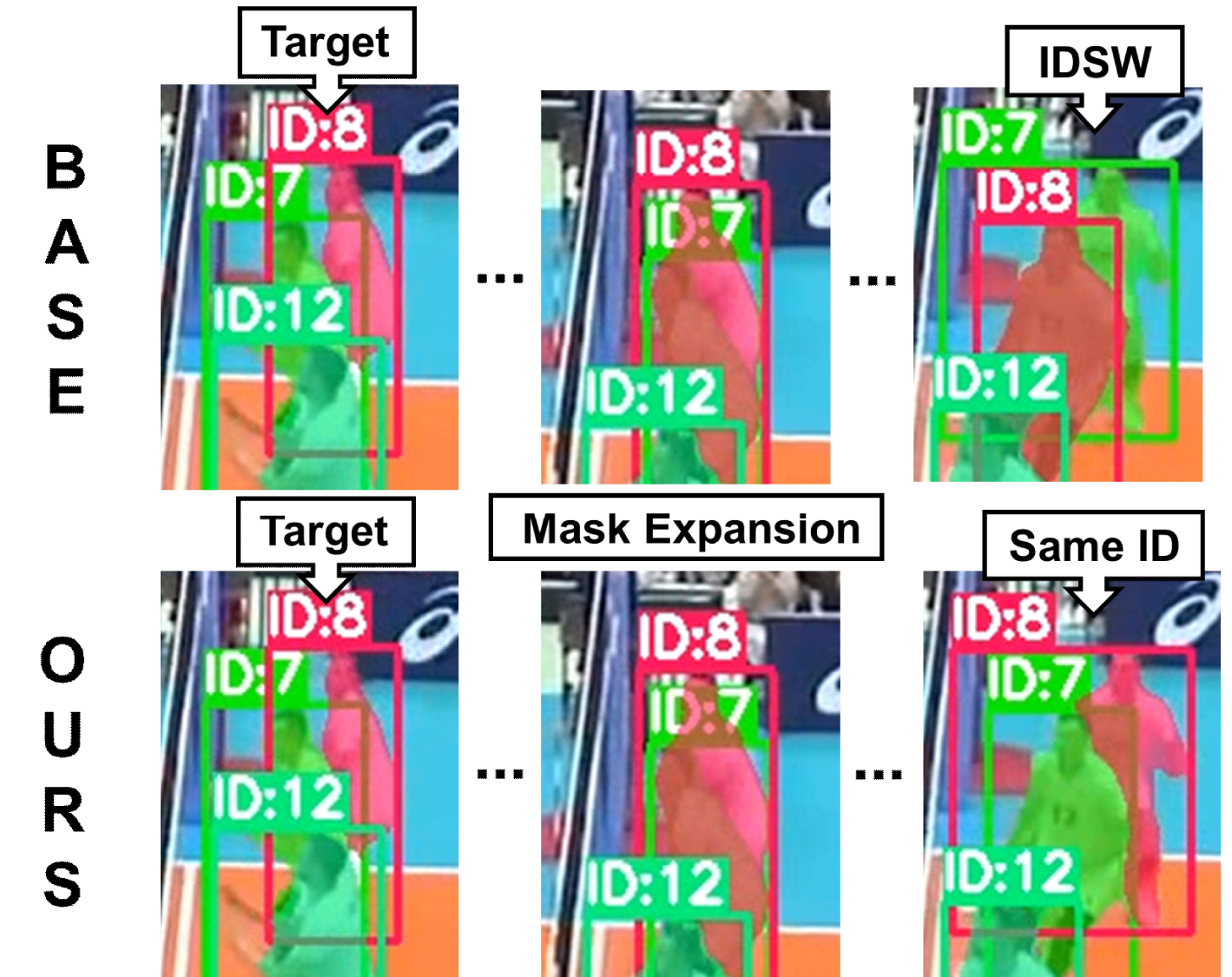}
        \caption{Scene 2: Prevention of ID switches enabled by H-CoI.}
        \label{subfig:result2}
    \end{subfigure}
    
    \vspace{1em}
    
    \begin{subfigure}{\linewidth}
        \centering
        \includegraphics[width=\linewidth]{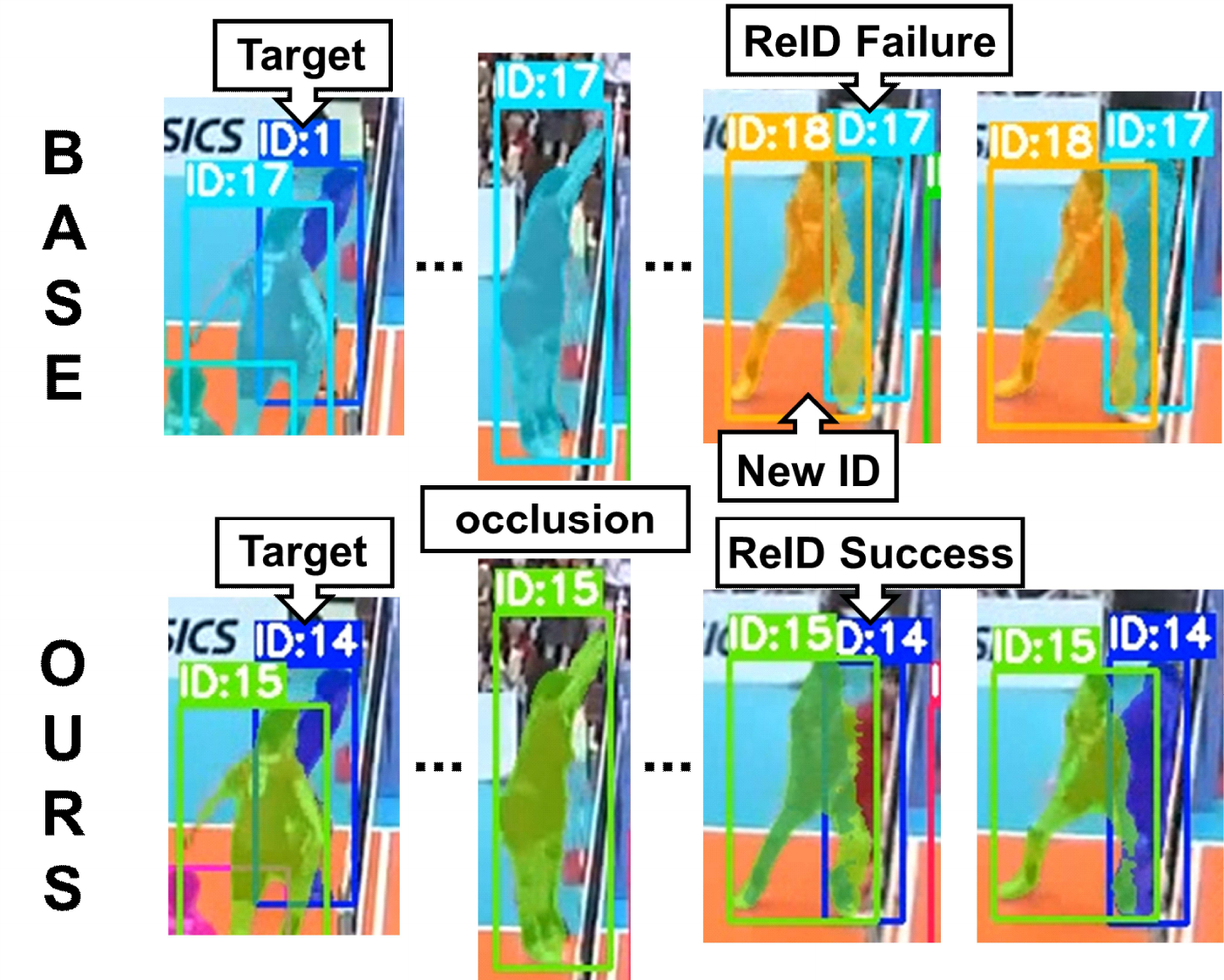}
        \caption{Scene 3: Recovery from corrupted masks enabled by Stage 2 matching.}
        \label{subfig:result3}
    \end{subfigure}
    
    \caption{Qualitative comparison in crowded sports scenarios included in the SportsMOT validation data.}
    \label{fig:track_result_list}
\end{figure}

Fig.~\ref{fig:track_result_list} shows the tracking results of SAM2MOT (“BASE”) and the proposed method (“OURS”) in three scenes (Fig.~\ref{subfig:result1},\ref{subfig:result2},\ref{subfig:result3}) included in the SportsMOT validation data.

\textbf{Scene 1 (Fig.~\ref{subfig:result1})} shows a sequence where the player with an orange mask performs a jump (left frame), undergoes occlusion by another player (center frame), and eventually reappears as the occlusion clears (right frame). In the baseline, SAM2MOT (top row), fails to associate the target with its original ID, erroneously assigning a new ID upon reappearance. This failure typically stems from the baseline's lack of adaptive mask control in crowded situations, which allows the features of the adjacent players to contaminate the target's feature during the overlap.

In contrast, SAMIDARE (bottom row) successfully maintains the target's ID through the DA-QR module. By evaluating local density, DA-QR suppresses improper mask re-generation in high-density regions. This prevents feature contamination from surrounding players and preserves the purity of the target’s discriminative features, enabling stable long-term tracking even during dense scenarios.

\textbf{Scene 2 (Fig.~\ref{subfig:result2})} illustrates a challenging scenario where a background player (pink mask in left frame) is partially overlapped by a foreground player (green mask in left frame). In the center frame, both the baseline and the proposed method exhibit mask expansion, particularly for the background player whose mask erroneously extends to the foreground player.

In the baseline (top row), an ID switch occurs in the right frame, where the pink mask is incorrectly transferred to the foreground player. This failure reveals a critical error in its corrupted mask handling: the baseline erroneously skipped the memory update for the foreground player (green mask), while incorrectly proceeding to update the memory for the background player (pink mask) despite its mask expansion. As a result, the background player's memory was updated with contaminated features from the foreground player, leading to ID switches.

In contrast, SAMIDARE (bottom row) maintains correct ID assignments. The H-CoI module enables this accurate ID maintenance by employing a hybrid reliability evaluation to differentiate between valid and degraded masks. It correctly identifies the expanded pink mask as unreliable and adaptively skips its memory update, while ensuring the valid features of the foreground player are preserved. By maintaining the purity of the stored features, SAMIDARE successfully prevents ID switches and recovers accurate segmentation masks as the players separate. This demonstrates the robustness of SAMIDARE in adaptively managing memory updates based on mask reliability, even in the presence of temporary segmentation errors.

\textbf{Scene 3 (Fig.~\ref{subfig:result3})} shows a four-frame sequence where the target player and a foreground player initially overlap (first frame). In the subsequent frames, both players perform a jump, causing the target to be occluded (second frame). By the final frames, the occlusion is resolved as they land, allowing the target to reappear.

In the baseline (top row), the foreground player's mask (light blue mask) erroneously shifts to the target player following the occlusion, and a new ID is assigned to the foreground player (third frame). This failure is attributed to the lack of adaptive mask control (DA-QR and H-CoI) under such dense conditions, which prevents the baseline from maintaining feature purity and generating accurate masks.

In contrast, SAMIDARE (bottom row) successfully maintains the correct ID for the foreground player (green mask). Notably, in the third frame, the target player’s mask (dark blue) initially shifts toward the foreground player as they separate. However, SAMIDARE achieves robust re-identification through Stage 2 matching in SA-OA, which leverages spatial information from the last frame where the target was successfully tracked prior to the occlusion. Crucially, as Stage 2 matching triggers the DA-QR process, SAMIDARE is able to re-generate a clean mask for the target in the fourth frame. 
\section{Conclusion}
\label{sec:conclusion}
We proposed SAMIDARE, a tracking-by-segmentation framework that enhances the baseline, SAM2MOT, for dense sports scenes through adaptive mask control (DA-QR, H-CoI) and state-aware new tracking ID assignment (SA-OA). By preserving feature purity and improving trajectory continuity for inactive tracks, SAMIDARE outperforms SOTA trackers on SportsMOT, particularly during frequent occlusion and frame-out events. These results confirm SAMIDARE as a comprehensive solution for sports tracking, effectively addressing the inherent limitations of mask-based methods in highly crowded scenarios.

Future work includes the dynamic adjustment of $\theta_{\text{density}}$. Since the optimal threshold varies by sport, automating this parameter based on player density and interaction patterns will be key to further enhancing generalizability across diverse sporting contexts.
{
    \small
    \bibliographystyle{ieeenat_fullname}
    \bibliography{main}

@String(CVPR= {IEEE Conf. Comput. Vis. Pattern Recog.})

@String(ICCV= {Int. Conf. Comput. Vis.})

@String(ECCV= {Eur. Conf. Comput. Vis.})

@String(ICLR = {Int. Conf. Learn. Represent.})

@String(CVPRW= {IEEE Conf. Comput. Vis. Pattern Recog. Worksh.})

@String(CVPR  = {CVPR})

@String(ICCV  = {ICCV})

@String(ECCV  = {ECCV})

@String(ICLR  = {ICLR})

@String(CVPRW= {CVPRW})

@article{DBLP:journals/corr/abs-2504-04519,
  author       = {Junjie Jiang and
                  Zelin Wang and
                  Manqi Zhao and
                  Yin Li and
                  Dongsheng Jiang},
  title        = {{SAM2MOT:} {A} Novel Paradigm of Multi-Object Tracking by Segmentation},
  journal      = {CoRR},
  volume       = {abs/2504.04519},
  year         = {2025},
}

@inproceedings{DBLP:conf/eccv/ZhangSJYWYLLW22,
  author       = {Yifu Zhang and
                  Peize Sun and
                  Yi Jiang and
                  Dongdong Yu and
                  Fucheng Weng and
                  Zehuan Yuan and
                  Ping Luo and
                  Wenyu Liu and
                  Xinggang Wang},
  title        = {ByteTrack: Multi-object Tracking by Associating Every Detection Box},
  booktitle    = ECCV,
  pages        = {1--21},
  year         = {2022},
}

@inproceedings{DBLP:conf/iccv/CuiZZYWW23,
  author       = {Yutao Cui and
                  Chenkai Zeng and
                  Xiaoyu Zhao and
                  Yichun Yang and
                  Gangshan Wu and
                  Limin Wang},
  title        = {SportsMOT: {A} Large Multi-Object Tracking Dataset in Multiple Sports
                  Scenes},
  booktitle    = ICCV,
  pages        = {9887--9897},
  year         = {2023},
}

@inproceedings{DBLP:conf/cvpr/CaoPWKK23,
  author       = {Jinkun Cao and
                  Jiangmiao Pang and
                  Xinshuo Weng and
                  Rawal Khirodkar and
                  Kris Kitani},
  title        = {Observation-Centric {SORT:} Rethinking {SORT} for Robust Multi-Object
                  Tracking},
  booktitle    = CVPR,
  pages        = {9686--9696},
  year         = {2023},
}

@inproceedings{DBLP:conf/cvpr/Qin0ZF0024,
  author       = {Zheng Qin and
                  Le Wang and
                  Sanping Zhou and
                  Panpan Fu and
                  Gang Hua and
                  Wei Tang},
  title        = {Towards Generalizable Multi-Object Tracking},
  booktitle    = CVPR,
  pages        = {18995--19004},
  year         = {2024},
}

@inproceedings{DBLP:conf/wacv/HuangYSKKLHH24,
  author       = {Hsiang{-}Wei Huang and
                  Cheng{-}Yen Yang and
                  Jiacheng Sun and
                  Pyong{-}Kun Kim and
                  Kwang{-}Ju Kim and
                  Kyoungoh Lee and
                  Chung{-}I Huang and
                  Jenq{-}Neng Hwang},
  title        = {Iterative Scale-Up ExpansionIoU and Deep Features Association for
                  Multi-Object Tracking in Sports},
  booktitle    = {WACV Workshop},
  pages        = {163--172},
  year         = {2024},
}

@inproceedings{DBLP:conf/cvpr/LvHZLH024,
  author       = {Weiyi Lv and
                  Yuhang Huang and
                  Ning Zhang and
                  Ruei{-}Sung Lin and
                  Mei Han and
                  Dan Zeng},
  title        = {DiffMOT: {A} Real-time Diffusion-based Multiple Object Tracker with
                  Non-linear Prediction},
  booktitle    = CVPR,
  pages        = {19321--19330},
  year         = {2024},
}

@inproceedings{DBLP:conf/cvpr/StanczykYB25,
  author       = {Tomasz Stanczyk and
                  Seongro Yoon and
                  Fran{\c{c}}ois Br{\'{e}}mond},
  title        = {No Train Yet Gain: Towards Generic Multi-Object Tracking in Sports
                  and Beyond},
  booktitle    = CVPRW,
  pages        = {6039--6048},
  year         = {2025},
}

@inproceedings{DBLP:conf/cvpr/LiKDPSG024,
  author       = {Siyuan Li and
                  Lei Ke and
                  Martin Danelljan and
                  Luigi Piccinelli and
                  Mattia Seg{\`{u}} and
                  Luc Van Gool and
                  Fisher Yu},
  title        = {Matching Anything by Segmenting Anything},
  booktitle    = CVPR,
  pages        = {18963--18973},
  year         = {2024},
}

@inproceedings{DBLP:conf/eccv/LiuZRLZYJLYSZZ24,
  author       = {Shilong Liu and
                  Zhaoyang Zeng and
                  Tianhe Ren and
                  Feng Li and
                  Hao Zhang and
                  Jie Yang and
                  Qing Jiang and
                  Chunyuan Li and
                  Jianwei Yang and
                  Hang Su and
                  Jun Zhu and
                  Lei Zhang},
  title        = {Grounding {DINO:} Marrying {DINO} with Grounded Pre-training for Open-Set
                  Object Detection},
  booktitle    = ECCV,
  pages        = {38--55},
  year         = {2024},
}

@inproceedings{DBLP:conf/iccv/KirillovMRMRGXW23,
  author       = {Alexander Kirillov and
                  Eric Mintun and
                  Nikhila Ravi and
                  Hanzi Mao and
                  Chlo{\'{e}} Rolland and
                  Laura Gustafson and
                  Tete Xiao and
                  Spencer Whitehead and
                  Alexander C. Berg and
                  Wan{-}Yen Lo and
                  Piotr Doll{\'{a}}r and
                  Ross B. Girshick},
  title        = {Segment Anything},
  booktitle    = ICCV,
  pages        = {3992--4003},
  year         = {2023},
}

@inproceedings{DBLP:conf/iccv/ChengOPSL23,
  author       = {Ho Kei Cheng and
                  Seoung Wug Oh and
                  Brian L. Price and
                  Alexander G. Schwing and
                  Joon{-}Young Lee},
  title        = {Tracking Anything with Decoupled Video Segmentation},
  booktitle    = ICCV,
  pages        = {1316--1326},
  year         = {2023},
}

@inproceedings{DBLP:conf/iclr/RaviGHHR0KRRGMP25,
  author       = {Nikhila Ravi and
                  Valentin Gabeur and
                  Yuan{-}Ting Hu and
                  Ronghang Hu and
                  Chaitanya Ryali and
                  Tengyu Ma and
                  Haitham Khedr and
                  Roman R{\"{a}}dle and
                  Chlo{\'{e}} Rolland and
                  Laura Gustafson and
                  Eric Mintun and
                  Junting Pan and
                  Kalyan Vasudev Alwala and
                  Nicolas Carion and
                  Chao{-}Yuan Wu and
                  Ross B. Girshick and
                  Piotr Doll{\'{a}}r and
                  Christoph Feichtenhofer},
  title        = {{SAM} 2: Segment Anything in Images and Videos},
  booktitle    = ICLR,
  year         = {2025},
}

@article{DBLP:journals/corr/abs-2107-08430,
  author       = {Zheng Ge and
                  Songtao Liu and
                  Feng Wang and
                  Zeming Li and
                  Jian Sun},
  title        = {{YOLOX:} Exceeding {YOLO} Series in 2021},
  journal      = {CoRR},
  year         = {2021},
}

@article{DBLP:journals/ijcv/LuitenODTGLL21,
  author       = {Jonathon Luiten and
                  Aljosa Osep and
                  Patrick Dendorfer and
                  Philip H. S. Torr and
                  Andreas Geiger and
                  Laura Leal{-}Taix{\'{e}} and
                  Bastian Leibe},
  title        = {{HOTA:} {A} Higher Order Metric for Evaluating Multi-object Tracking},
  journal      = {Int. J. Comput. Vis.},
  volume       = {129},
  number       = {2},
  pages        = {548--578},
  year         = {2021},
}

@inproceedings{DBLP:conf/eccv/RistaniSZCT16,
  author       = {Ergys Ristani and
                  Francesco Solera and
                  Roger S. Zou and
                  Rita Cucchiara and
                  Carlo Tomasi},
  editor       = {Gang Hua and
                  Herv{\'{e}} J{\'{e}}gou},
  title        = {Performance Measures and a Data Set for Multi-target, Multi-camera
                  Tracking},
  booktitle    = ECCV,
  volume       = {9914},
  pages        = {17--35},
  year         = {2016},
}

@article{DBLP:journals/bmcbi/VeeramaniRC18,
  author       = {Balaji Veeramani and
                  John W. Raymond and
                  Pritam Chanda},
  title        = {DeepSort: deep convolutional networks for sorting haploid maize seeds},
  journal      = {{BMC} Bioinform.},
  volume       = {19-S},
  number       = {9},
  pages        = {85--93},
  year         = {2018},
}

@article{DBLP:journals/tmm/DuZSZSGM23,
  author       = {Yunhao Du and
                  Zhicheng Zhao and
                  Yang Song and
                  Yanyun Zhao and
                  Fei Su and
                  Tao Gong and
                  Hongying Meng},
  title        = {StrongSORT: Make DeepSORT Great Again},
  journal      = {{IEEE} Trans. Multim.},
  volume       = {25},
  pages        = {8725--8737},
  year         = {2023},
}

@article{DBLP:journals/corr/abs-2206-14651,
  author       = {Nir Aharon and
                  Roy Orfaig and
                  Ben{-}Zion Bobrovsky},
  title        = {BoT-SORT: Robust Associations Multi-Pedestrian Tracking},
  journal      = {CoRR},
  volume       = {abs/2206.14651},
  year         = {2022},
}

@inproceedings{DBLP:conf/cvpr/QinZ0D0T23,
  author       = {Zheng Qin and
                  Sanping Zhou and
                  Le Wang and
                  Jinghai Duan and
                  Gang Hua and
                  Wei Tang},
  title        = {MotionTrack: Learning Robust Short-Term and Long-Term Motions for
                  Multi-Object Tracking},
  booktitle    = CVPR,
  pages        = {17939--17948},
  year         = {2023},
}

@article{DBLP:journals/tip/YangHCJH26,
  author       = {Cheng{-}Yeng Yang and
                  Hsiang{-}Wei Huang and
                  Wenhao Chai and
                  Zhongyu Jiang and
                  Jenq{-}Neng Hwang},
  title        = {{SAMURAI:} Motion-Aware Memory for Training-Free Visual Object Tracking
                  With {SAM} 2},
  journal      = {{IEEE} Trans. Image Process.},
  volume       = {35},
  pages        = {970--982},
  year         = {2026},
}

@article{DBLP:journals/corr/abs-2410-16268,
  author       = {Shuangrui Ding and
                  Rui Qian and
                  Xiaoyi Dong and
                  Pan Zhang and
                  Yuhang Zang and
                  Yuhang Cao and
                  Yuwei Guo and
                  Dahua Lin and
                  Jiaqi Wang},
  title        = {SAM2Long: Enhancing {SAM} 2 for Long Video Segmentation with a Training-Free
                  Memory Tree},
  journal      = {CoRR},
  volume       = {abs/2410.16268},
  year         = {2024},
}

@misc{sam2github,
  author = {Nikhila Ravi and
              Valentin Gabeur and
              Yuan{-}Ting Hu and
              Ronghang Hu and
              Chaitanya Ryali and
              Tengyu Ma and
              Haitham Khedr and
              Roman R{\"{a}}dle and
              Chlo{\'{e}} Rolland and
              Laura Gustafson and
              Eric Mintun and
              Junting Pan and
              Kalyan Vasudev Alwala and
              Nicolas Carion and
              Chao{-}Yuan Wu and
              Ross B. Girshick and
              Piotr Doll{\'{a}}r and
              Christoph Feichtenhofer},
  title = {Segment Anything Model 2.1},
  howpublished = {\url{https://github.com/facebookresearch/sam2?tab=readme}},
  year = {2024},
  note = {Accessed: 2026-03-05},
}
}


\end{document}